# Orange Lab: Lowering Barriers to Data Mining through Embedded Interactive Workflows

Matej Bevec
University of Ljubljana
Ljubljana, Slovenia
matej.bevec@fri.uni-lj.si

Aleš Erjavec
University of Ljubljana
Ljubljana, Slovenia
ales.erjavec@fri.uni-lj.si

Vesna Tanko
Revelo d.o.o.
Ljubljana, Slovenia
vesna@revelo.bi

Lena Trnovec
University of Ljubljana
Ljubljana, Slovenia
lena.trnovec@fri.uni-lj.si

Lan Žagar
Revelo d.o.o.
Ljubljana, Slovenia
lan@revelo.bi

Ana Farič
University of Ljubljana
Ljubljana, Slovenia
ana.faric@fri.uni-lj.si

Janez Demšar
University of Ljubljana
Ljubljana, Slovenia
janez.demsar@fri.uni-lj.si

Blaž Zupan
University of Ljubljana
Ljubljana, Slovenia
blaz.zupan@fri.uni-lj.si

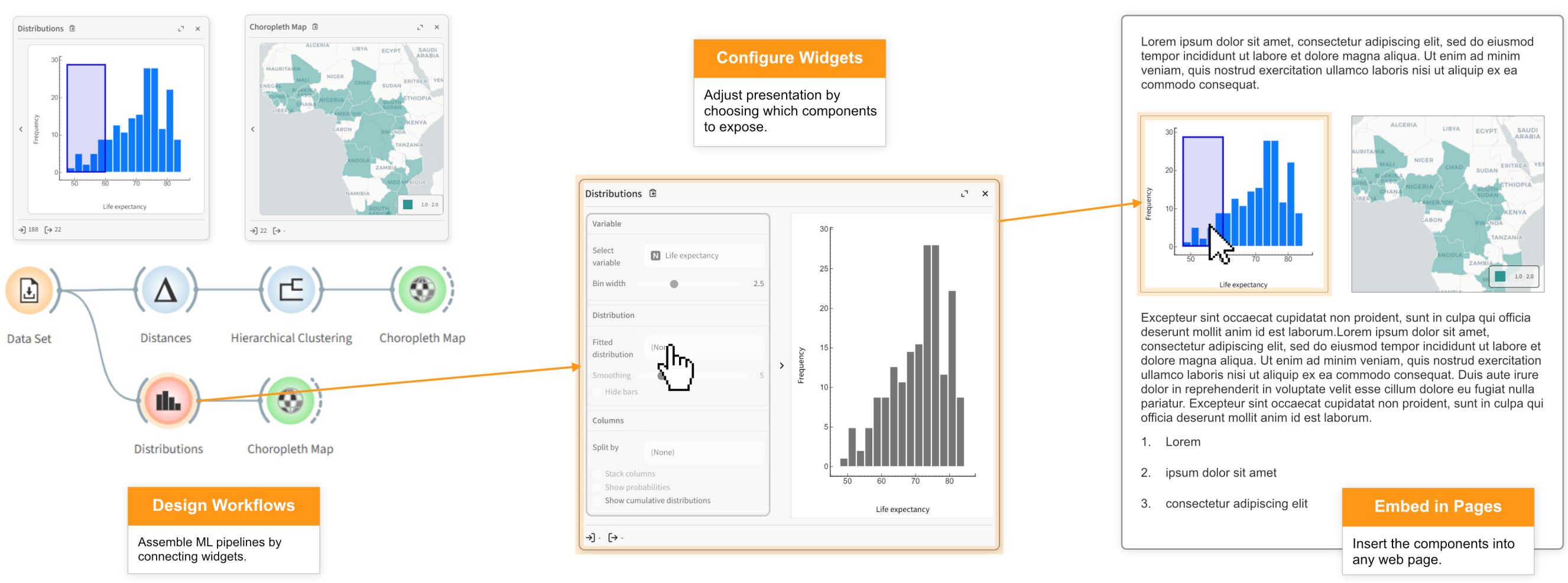

**Figure 1: Widget exposition in Orange Lab. Users construct data mining workflows through visual programming on a collaborative canvas. Widgets can then be simplified by choosing which of their components to expose or hide. Finally, the configured widgets are embedded into any web page to create interactive interfaces. Embedded widgets remain part of of pipeline – interacting with one will be reflected in others.**

## Abstract

While visual programming of data analysis workflows has become an important vehicle for the democratization of data science, such systems remain largely confined to standalone applications and offer limited support for transitioning their visual analytics solutions into interactive web environments. As a result, data analysis pipelines are difficult to share, embed, and adapt into user-facing analytical tools. We present Orange Lab, a web-based collaborative environment for visual data analytics. At its core, Orange Lab enables users to visually construct machine learning workflows from modular components, where interactions in any component propagate seamlessly through the workflow, turning static pipelines into dynamic, reactive systems that support exploration and data-driven storytelling. Our key contribution is component exposition, a paradigm that allows authors to embed selected workflow components, or parts of their interfaces, into arbitrary web contexts, creating synchronized, interactive interfaces while hiding underlying workflow complexity. This enables the development of tailored analytical views and narrative-driven experiences that integrate data analysis directly into online materials. We demonstrate the approach through deployments in data literacy education, where embedded components guide students in hands-on exploration of machine learning concepts without requiring knowledge of the underlying system, showing that Orange Lab effectively lowers barriers to entry and supports the democratization of data science.

# 1 Introduction

Accessible data mining and machine learning (ML) have become increasingly important across a wide range of domains, including the natural and social sciences, business, and organizational decision-making. The ability to extract insights from raw data is now a critical skill, yet many people who would benefit from these techniques, such as domain experts, educators, and decision makers, do not have formal programming experience. Enabling these users to engage with data-driven methods remains an open challenge.

*Visual programming* has emerged as a successful approach to democratizing data science. In particular, the *dataflow* paradigm allows users to construct analytical pipelines by connecting prebuilt components into computational graphs on a visual canvas. By interacting with graphical user interfaces rather than writing code, users can iteratively design, modify, and execute complex workflows. A number of systems follow this philosophy, including KNIME [2], RapidMiner [15, 17], VisFlow [29], WEKA [14] and our predecessor, Orange [11], along with various business-oriented analytics platforms, like SPSS Modeler [28], SAS Enterprise Miner [19], Alteryx Designer [1] or Dataiku DSS [10]. Some of these tools, such as Orange, further emphasize interactivity and enable users to manipulate data and immediately observe how results and visualizations update in real time. This tight feedback loop supports not only ease of use for non-programmers, but also rapid prototyping, exploratory analysis, and rich interactive visualization, which are often more difficult to achieve in code.

Despite these advantages, a key limitation remains. Even though visual programming lowers the barrier to entry, it still requires users to learn the interface to the tool itself [22]. In many scenarios, such as teaching data mining concepts to students or presenting results to stakeholders, this requirement is undesirable. Instructors and presenters often wish to focus on domain concepts rather than on the mechanics of a specific tool. Moreover, even when teaching the tool is within scope, there is frequently a need to guide users through a curated sequence of interactions and insights. This is particularly relevant in web-based lessons, reports, or dashboards, where interactive elements should be tightly coupled with explanatory content. Switching between a standalone tool and accompanying material introduces unnecessary cognitive load and distracts users with interface elements and functionality that are irrelevant to the task at hand.

Current visual programming tools do not yet address this issue. While they enable the construction of powerful workflows, they do not allow individual components to be easily embedded into external environments, such as web pages, or not in a way that preserves interactivity. As a result, workflows are difficult to share, adapt, and integrate into user-facing analytical tools or narratives.

To address this gap, we introduce *Orange Lab*, which offers the standard dataflow workflow construction model, familiar to Orange users, but now, implemented as a web app, with additional support for real-time collaboration. Crucially, though, Orange Lab introduces a novel *component exposition* paradigm, which allows authors to extract the individual components (called widgets) and expose their interactive interfaces in arbitrary web environments. These embedded components remain fully interactive and synchronized with the underlying workflow, while hiding the complexity of the full data analysis pipeline from the user. The paradigm goes even beyond exposing selected components: it supports exposing selected parts of the component's interface, allowing the app designers to guide users' attention while minimizing distractions. Such flexibility in choosing the functionality and interface to be incorporated within web pages supports a range of applications, including interactive educational materials that encourage exploratory learning. Component exposition enables the creation of tightly integrated, interactive narratives in which visualization and user interaction are seamlessly combined with instructional or explanatory content.

In summary, our contributions are threefold: (1) we present Orange Lab, an open-source web-based collaborative toolkit for data mining through visual programming; (2) we introduce the component exposition paradigm, enabling the embedding of customized, interactive workflow components into arbitrary web contexts; and (3) we demonstrate the effectiveness of this approach through real-world deployments in data literacy education.

# 2 Related Work

The design of Orange Lab is related to the areas of visual dataflow programming environments, systems for building interactive machine learning applications, and methods for composing interactive web-based interfaces. Each of these approaches contributes to making data analysis more accessible and interactive, but have rarely been considered together. As we discuss them, we highlight the gap between visual workflow programming and the creation of embeddable, user-facing analytical interfaces.

## 2.1 Dataflow programming tools

Visual dataflow programming environments have played a central role in making data science more accessible by allowing users to construct workflows by connecting modular components. These systems all offer some level of data analysis, visualization, and machine learning capabilities, but differ in their target audiences and design philosophies.

Tools such as Orange [11], WEKA Knowledge Flow [14], and VisFlow [29] emphasize machine learning and exploratory analysis. Orange, in particular, focuses on interactive visualization, rapid prototyping and accessibility to new users. WEKA Knowledge Flow provides a conceptually similar interface grounded in an extensive traditional machine learning repertoire, though with more limited visualization capabilities and a less modern user experience. VisFlow adopts a comparable interface to Orange, with reactive visualizations and an entirely canvas-based interface, but is a smaller project with a limited set of components and configuration options.

At the other end of the spectrum, platforms such as Altair RapidMiner [15, 17] and KNIME [2] target more advanced and enterprise-oriented use cases. These systems support the construction of end-to-end data pipelines, including deployment and automation. KNIME occupies a middle ground, offering a broad ecosystem that spans research and enterprise applications, though with a more technical interface and less emphasis on dynamic, exploratory visualization. RapidMiner, similarly, provides a comprehensive suite with growing support for modern machine learning and AI capabilities, but again prioritizes production readiness over accessibility.

A broader class of dataflow-based systems focuses on business analytics and data engineering, including IBM SPSS Modeler [28], SAS Enterprise Miner [19], Alteryx Designer [1], Dataiku DSS [10], and Azure Machine Learning Designer [20]. These platforms are centered around organizational workflows involving statistics, ETL processes, and ML Ops. However, they typically place less emphasis on interactive visual analytics and exploratory workflows.

Some of these toolkits provide functionality that is particularly relevant to our work, although with important limitations. Orange, WEKA, SPSS Modeler, SAS Enterprise Miner, and Alteryx Designer are primarily desktop-based and do not offer web interfaces. In contrast, VisFlow, RapidMiner, KNIME, Dataiku DSS and Azure ML Designer do provide web-based access, with all but VisFlow additionally supporting some level of collaboration, though only Dataiku offers real-time shared editing on the canvas. Support for dashboard-like user interfaces varies significantly. Orange, WEKA, VisFlow, SPSS Modeler, and SAS Enterprise Miner are restricted to static reporting and do not enable the construction of interactive dashboards. RapidMiner, KNIME, Dataiku DSS and Azure ML Designer on the other hand, offer more advanced dashboarding capabilities – particularly KNIME, with its so-called "data apps".

Despite their strengths, most of these systems remain largely confined to their native ecosystems. While some provide dashboarding support, they generally do not support embedding workflow components into arbitrary web contexts, limiting their use in narrative-driven or educational settings.

## 2.2 Interactive ML systems and dashboards

A related line of work focuses on tools for building interactive machine learning systems and user-facing dashboards. These systems enable the composition of interactive components into applications, but are typically accessed through programming rather than visual interfaces.

Marcelle [12] is a system which provides high-level abstractions for constructing both machine learning workflows and corresponding user-facing graphical dashboards through a unified JavaScript library. While this lowers the barrier to development compared to low-level tools, it still requires programming expertise. In contrast, our approach enables interface construction through direct manipulation of existing components, allowing authors to hide select elements with a few clicks. In this sense, Marcelle is additive and code-driven whereas our system is subtractive and GUI-based.

Other libraries, such as MLJ [4] and Metaflow [13], also provide a high-level interface to creating machine learning pipelines but do not address the construction of interactive interfaces. Conversely, tools like Lobe [21] and Teachable Machine [7] provide fully integrated, user-friendly interfaces for training models (e.g. by simply uploading training samples), and have been used particularly in educational or entertainment contexts. However, these systems are limited to a fixed workflow within a specific domain (e.g. image and video), offering little flexibility or extensibility.

Business intelligence platforms, including Tableau [24], Sisense [26], and Embeddable [18], also support the creation of interactive dashboards by combining programmatic APIs with no-code tools. While these systems are effective for delivering polished, user-facing analytics, the resulting interfaces are typically confined within their respective ecosystems and are not easily embedded into arbitrary web contexts.

## 2.3 Composable web interfaces

Finally, there exist systems for integrating computation with narrative content to create interactive, web-based materials. A prominent example is the ecosystem surrounding Jupyter Notebooks [16], which combine executable code, text, and visual outputs in a single document format. This paradigm supports the creation of narratives, such as interactive tutorials, lessons, or demos.

Platforms such as Google Colab [3] and CoCalc [23] extend this model to the cloud, allowing users to build, share and run notebooks on the web. CoCalc, in particular, is designed around collaboration, offering live cursors, chat and other features. With its course management tools it also places a strong emphasis on educational use.

While these services remain largely self-contained, creators who require interactivity in their own web materials usually turn to visualization libraries such as web-native D3.js [6] or Vega-Lite [25], or Python-based libraries like bqplot [5]. While flexible, these libraries are low-level tools for developers, and still require significant effort to implement the underlying data processing and coordination between visualized elements.

Recent tools such as Voilà [9] and Mercury [27] attempt to bridge this gap by transforming notebooks into interactive web applications. They let users create notebooks combining GUI elements from libraries like ipywidgets [8] and visualizations from libraries like bgplot. The code blocks can then be hidden, leaving an interactive web page where visualizations can be controlled through GUI elements.

Such frameworks are designed to share notebooks with non-technical users. Although these interfaces are still confined to the system at large (cannot be embedded anywhere), the support for text-based accompanying material means they are nonetheless a good fit for creating interactive narratives. However, they do not cover the underlying data processing pipeline, which still has to be devised in code through a lot of manual engineering, especially if rich reactivity and dynamic visualization are desired.

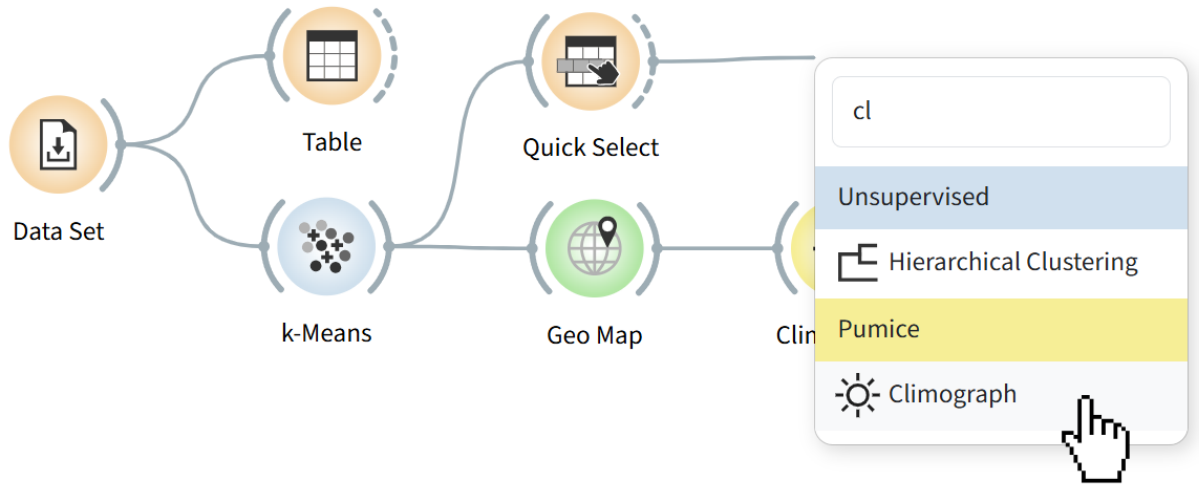

**Figure 2: Workflow construction. Users connect components called widgets to perform data analysis. Here, clustering is used to identify climate types based on monthly temperature and precipitation.**

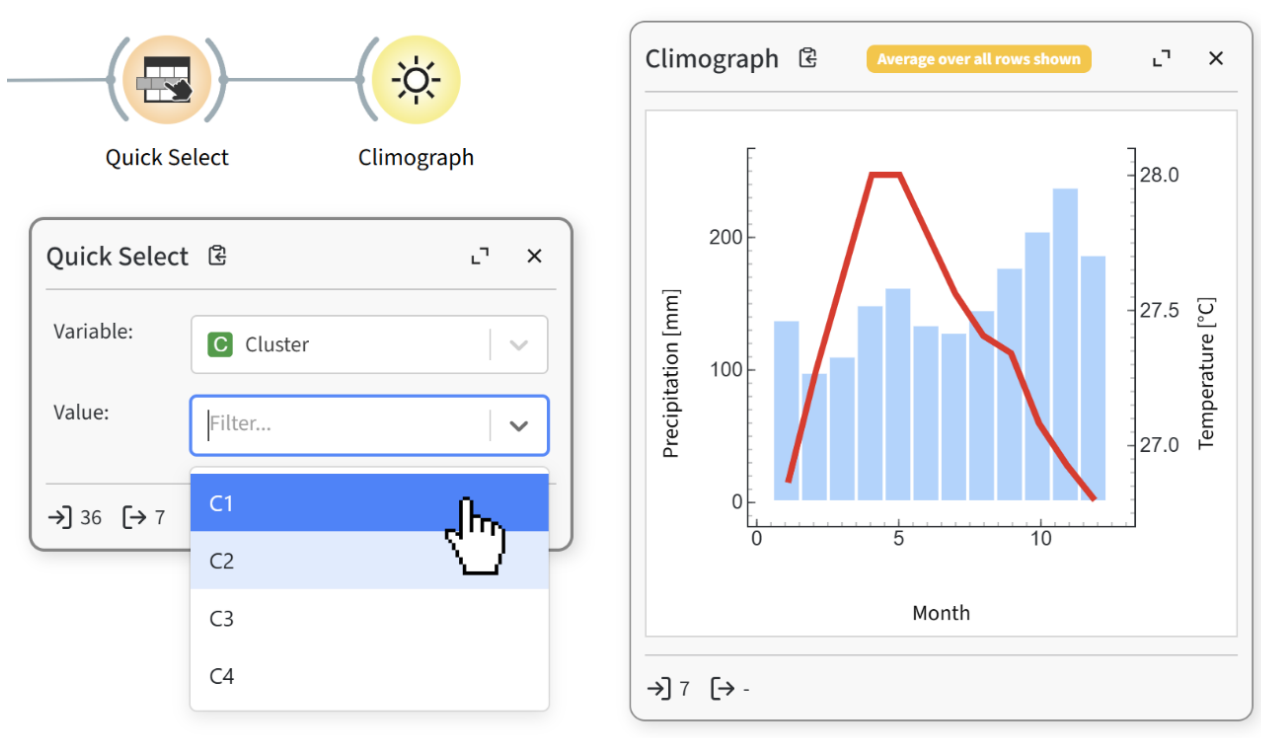

**Figure 3: Interacting with widgets. Users change settings and see downstream widgets react. Here, a particular climate cluster is selected to be displayed in the Climograph.**

## 3 Design Principles

We describe the key design principles underlying Orange Lab. These principles build on the foundations of visual dataflow programming while extending them to support collaborative use and integration with web-based environments. In particular, we focus on how workflows can be transformed from standalone visual analytics pipelines into interactive web apps, and how their components can be selectively exposed to create tailored user-facing interfaces.

### 3.1 Interactive Analysis through Visual Programming

The core design principle of Orange Lab, and its desktop-based predecessor, Orange, is in using visual programming to enable domain experts, business executives, educators and other non-programmers to construct data analysis and visualization pipelines without the need to code, as well as to offer individuals who may be skilled programmers to prototype solutions faster and communicate their ideas about data better and more intuitively.

The main building block in Orange Lab is a widget, a modularized, independent component which takes in some inputs, computes some analysis or visualization, and produces appropriate outputs. The widget catalogue is composed so that each widget performs a consistent amount of work (roughly one data analysis task or operation) and is consistently named and categorized according to its task. This is with a goal to create an interaction language that can become intuitive to the user and allow them to anticipate what widget they might require next.

Users connect such widgets on an interactive canvas to build data mining workflows (Figure 2). In a connected workflow, users can adjust their analysis by interacting with widgets and configuring their operation by changing their so-called settings. These changes propagate through the network to all downstream widgets, creating a system that is reactive to user input (Figure 3).

This reactivity, along with another key feature – interactive and dynamic visualizations – means Orange Lab lends itself not only to data processing, but also to so-called exploratory data analysis. Since all visualizations are designed to adapt to incoming data and are in some way selectable (and propagate their selections downstream), users can “click around” the data and see their analysis adapt, exposing patterns, providing intuition behind data analysis, helping domain experts communicate etc.

This design is convenient not only for users, but also to developers. Because widgets are fully encapsulated, i.e. entirely defined by their inputs and settings and not affected by any external state, they are also very easy to develop for external collaborators, even those new to the project, who don’t need to understand any of the overarching systems to add new features. Furthermore, this means that any potential bad practices in widget development will not compound in the system at large. As such, and due to the fact that widgets all share a consistent design API, they may also be a safe target for AI-assisted development.

### 3.2 Widget Exposition

Even though Orange Lab’s design principles make it friendly to developers and intuitive for new users to learn, there are, as described in Section 1, situations where we would like to make it accessible in an even simpler format to individuals who are not familiar with the application.

To address this, we introduce the widget exposition system. Orange Lab widgets can be injected into any web page by obtaining a special link and embedding it within an iFrame element (Figure 4). This allows the author to take the relevant widgets and include them anywhere where they could benefit from offering interactive visualization or analysis. The viewers of the resulting page do not need to be aware of the underlying workflow, or even of Orange Lab, the toolkit as a whole. Yet, the embedded widgets remain connected as part of the pipeline, meaning interacting with one will be reflected in others, as one would expect in the classic dataflow environment. This facilitates the creation of pages which are interactive not only within each widget, but across the entire page (Figure 5).

Furthermore, when embedding a widget, the author can manipulate its appearance and simplify it by choosing to hide parts of the interface (e.g. individual settings, the plot, footer and header, etc.). This means they can tailor the visitors’ interaction with their activity exactly to their needs, such as by hiding information that may only confuse them and instead draw their attention to the relevant insights. This approach ensures that the creator is not limited by the interface defined by Orange, but can essentially build their own interfaces by combining multiple widgets, which proves convenient for various scenarios.

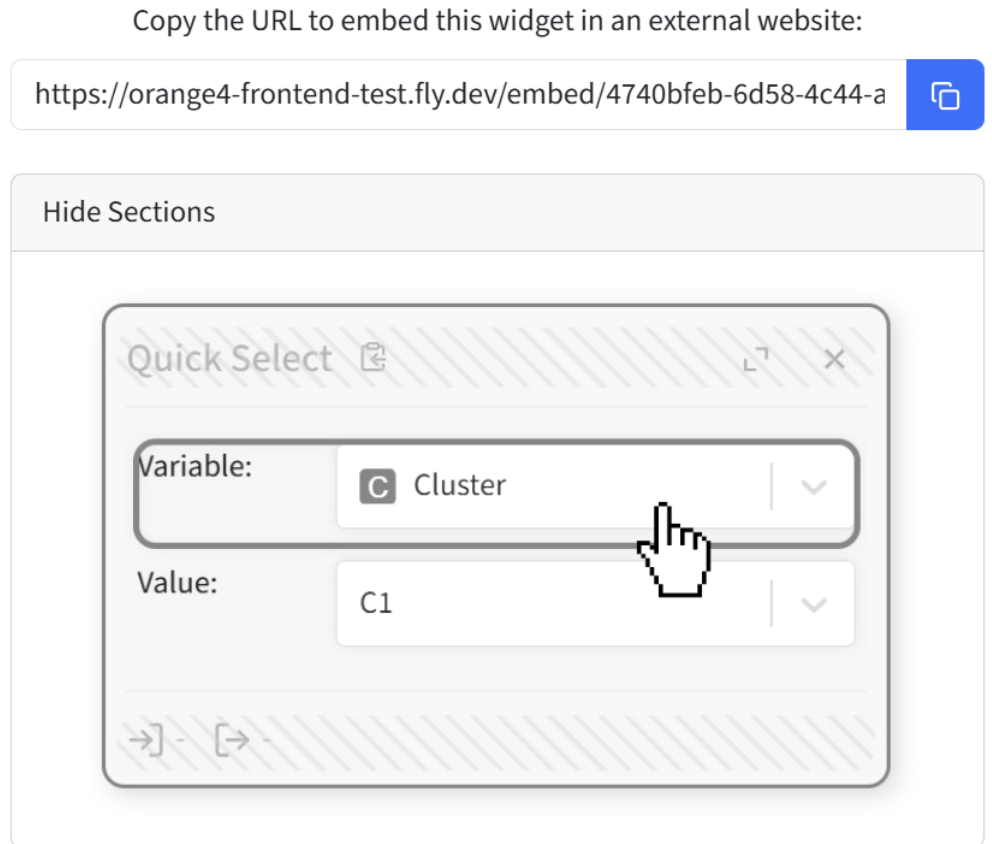

**Figure 4: Modifying a widget's appearance is straightforward: clicking on interface components hides or reveals them. A link to the widget – which also encodes the visibility of its components – can be copied for sharing. Here, we generate a link to embed a simplified widget for selecting clusters corresponding to climate zones.**

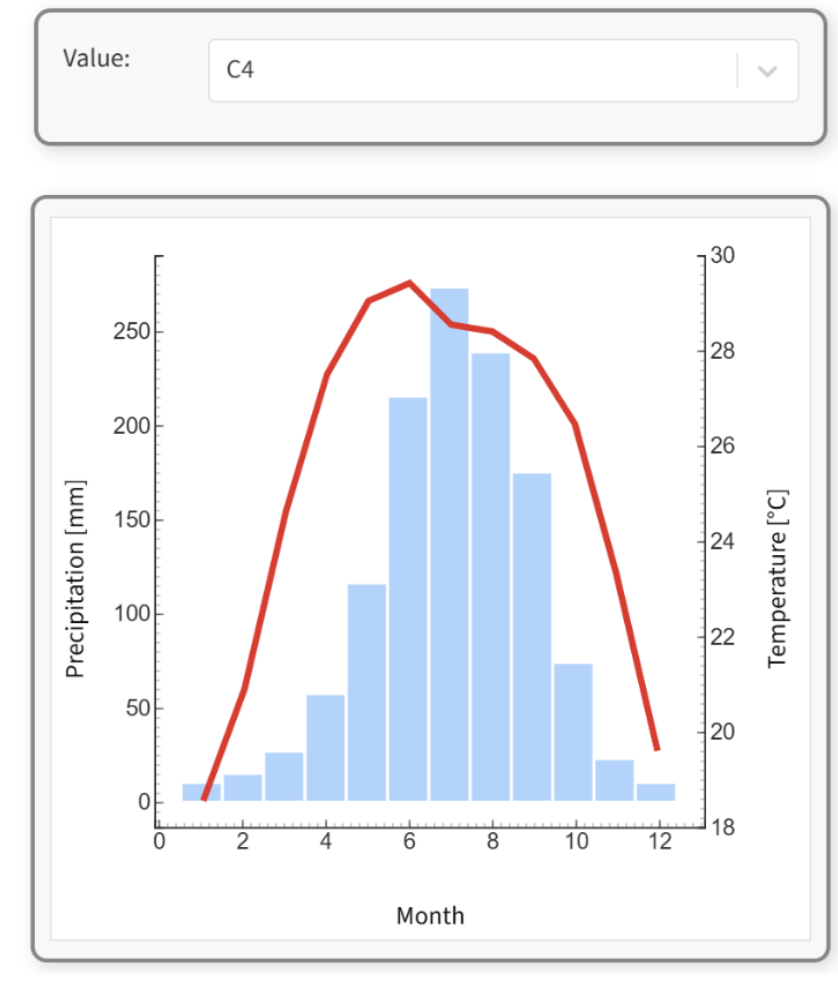

**Figure 5: Embedding widgets in a web page. Customized widgets can be injected into an arbitrary web page within iFrame elements. Here, only the interface components relevant to a lesson on climate are exposed.**

For example, one may wish to add some interactive visualization elements to a landing page for newly published research work, allowing the visitors to explore the results on the spot. Alternatively, we could construct a business intelligence pipeline, then create an automatically updating interactive dashboard as part of the company's internal web pages, allowing a manager to survey the relevant information. Finally, as described in Section 1, the solution is ideal for interactive lessons, quizzes and activities, where students can be guided to interact with data and reach conclusions on their own (Figure 5).

## 4 Implementation

We implement Orange Lab as a web-based system that facilitates the creation, within a collaborative client interface, of reactive workflows to be executed within a backend engine. Its architecture supports real-time interaction, multi-user synchronization, and integration of workflow elements into external web environments. The following describes the core system design and the mechanisms that enable these capabilities.

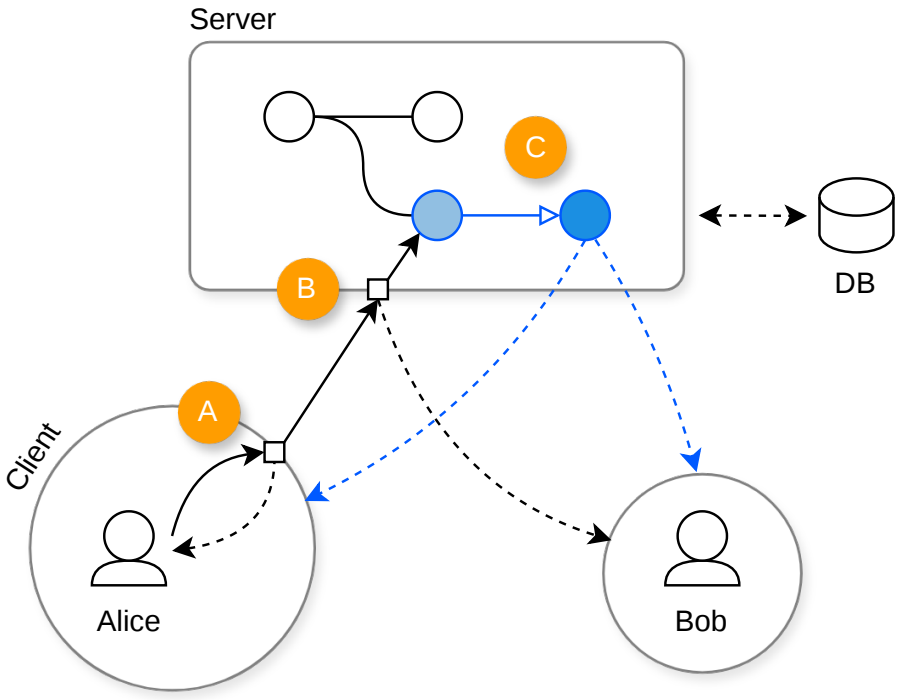

**Figure 6: Communication in collaborative use. When a user changes a setting, the change is immediately returned to reduce latency (A). On the server, the change is saved and broadcast to other users (B). Finally, the change triggers a recomputation of downstream nodes, producing results, which are broadcast to all users (C).**

### 4.1 Application Architecture

Orange Labs's application logic is built around a workflow: a computation graph composed of interconnected widget nodes. Each node consumes inputs and produces outputs, collectively forming a data mining pipeline that transforms raw data into results.

Each widget is configured through parameters called settings. When a setting changes, the widget recomputes as needed and emits updated outputs. These propagate through the workflow, triggering recomputation of downstream widgets. In this way, the workflow structure together with its settings defines the current state of the system.

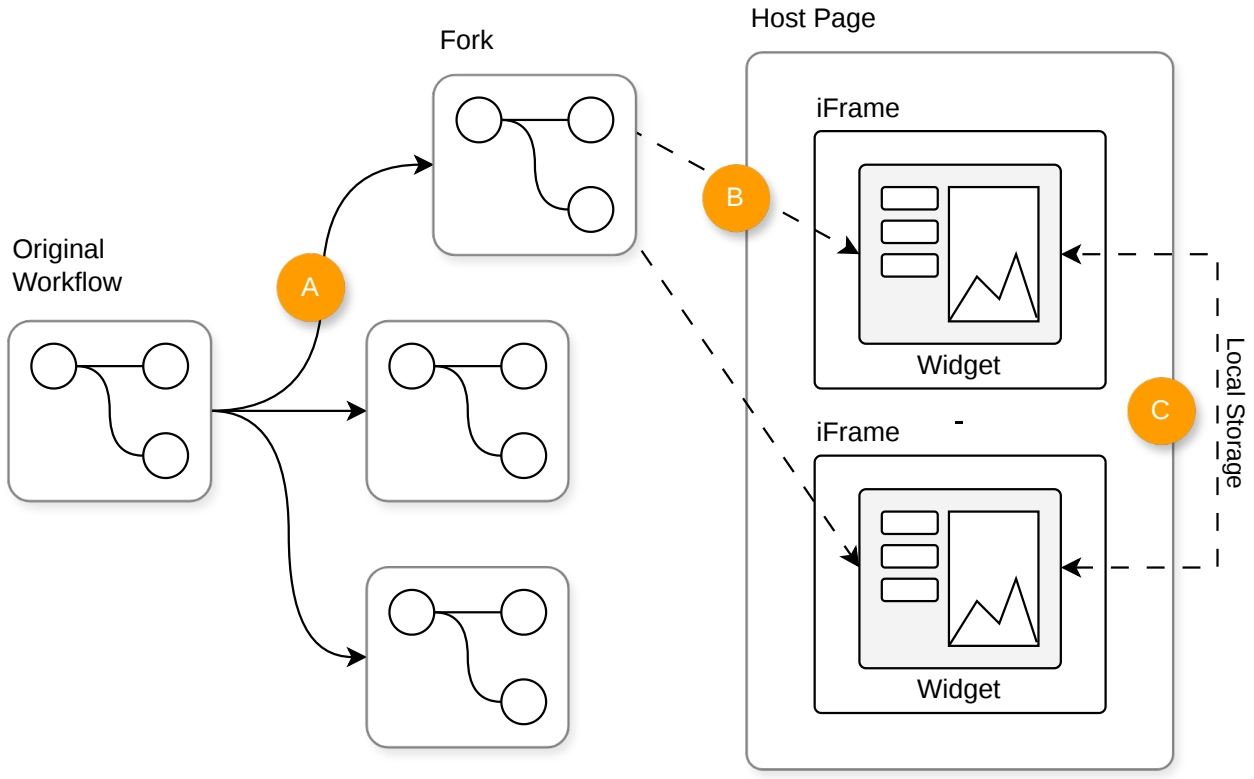

**Figure 7: The embedding system. When a viewer first accesses the host page, the original workflow is forked to create their own copy (A). During interaction, the embedded widgets communicate with the forked workflow (B). All widgets in a page coordinate via browser storage (C).**

This state is maintained on a Python backend, which serves multiple users connected via React clients. All computations are executed on the backend, while clients render a visual representation of the workflow and its widgets on an interactive canvas. Users can inspect the workflow, view computed visualizations, and interact with widgets to adjust their configuration. These changes are communicated to the backend over WebSockets by transmitting serialized (JSON) settings.

A typical interaction proceeds as follows (see Fig. 6). When a user modifies a setting, the change is immediately reflected locally in the client to ensure responsiveness. The update is then serialized and sent to the backend, where it is persisted in the database and broadcast to other connected clients. The backend applies the new setting to the corresponding widget, triggering recomputation of that node and any dependent nodes. Once computation completes, the resulting outputs (e.g., visualizations) are serialized and broadcast to all clients.

For performance reasons, the in-memory workflow is modified directly. To ensure consistency under multi-user access, operations are executed in an isolated, threaded manner, so that overlapping actions may invalidate one another, but the widgets are guaranteed to remain in a valid state. The backend as a whole is designed to scale across workflows. Incoming traffic can be distributed across multiple runner instances as long as users connected to the same workflow are routed to the same machine.

## 4.2 Widget Embedding System

While the collaborative workflow architecture shares many characteristics with existing dataflow systems, the widget exposition layer necessitates a more unconventional and novel approach.

Embedded widgets are, in essence, instances of the application running inside iframes, each configured to display a single widget with a selected subset of its interface elements exposed. This design allows them to be embedded into arbitrary web pages via generated embed links, which also encode the visibility of individual components in a format that is sufficiently expressive to remain human-readable and editable without reopening the graphical interface. From the developers' perspective, the exposition logic is encoded in reusable GUI components, meaning creators of new widgets do need to be aware of it.

Visitors to a web page should experience widgets exactly as in the original workflow and should be able to interact with them. However, because these interactions must not modify the source workflow, embedded widgets cannot connect to it directly. Instead, per-user forks are created dynamically: when a user first loads the page, a separate copy of the source workflow is instantiated and all embedded widgets are connected to this fork. The reference to the newly created fork is stored in the browser's session storage and reused on subsequent reloads, thereby preserving changes within the ongoing browser session (i.e. within the same tab). Opening the page in a new tab or window creates a fresh fork, although this persistence level can be adjusted – for example to extend beyond a single session.

A remaining challenge is coordinating interactions between widgets. As independent application instances, they are unaware of each other and appear to the backend as separate connections, which would normally result in multiple forks being created. To prevent this, local browser storage is used to coordinate the widgets, ensuring that a single fork is created and shared across all widgets on the page, allowing them to communicate as a unified system.

# 5 Case Studies

To demonstrate how Orange Lab – and, in particular, its widget exposition functionality – can improve accessibility to data science, we evaluate its use through large-scale, real-world deployments in education. The described activities focus on teaching data literacy to primary and secondary school students, introducing fundamental concepts from data mining and machine learning through hands-on exploration. We are currently testing this approach in over 70 schools across Slovenia, Luxembourg, Ireland, and Italy as part of a Europe-wide educational initiative.

Our initial approach was to present a – preferably real-world – dataset and use the desktop version of Orange to analyze it, either in a teacher-led or a hands-on setting. The primary advantage of Orange lies in its flexibility, which allows lessons to be adapted by exploring questions that arise during implementation. However, this approach revealed two key challenges. Most teachers were intimidated by the breadth of Orange's functionality and did not feel confident using it in the classroom. Preparing workflows in advance and giving teachers detailed instructions for using did not alleviate these fears. Students, in contrast, were not hesitant to explore the software; however, the lessons became overly focused on *how* to perform specific actions rather than on *what* was being done and why.

As most schools curricula do not include a dedicated subject on data literacy, we integrated these lessons into existing subjects, which introduced two additional challenges. Teachers from disciplines such as geography or biology were not prepared to invest time in learning specialized software for just a few lessons. Furthermore, relocating a single lesson to a computer lab with desktop machines proved impractical, while tablets – often more readily available in classrooms – cannot run software such as Orange.

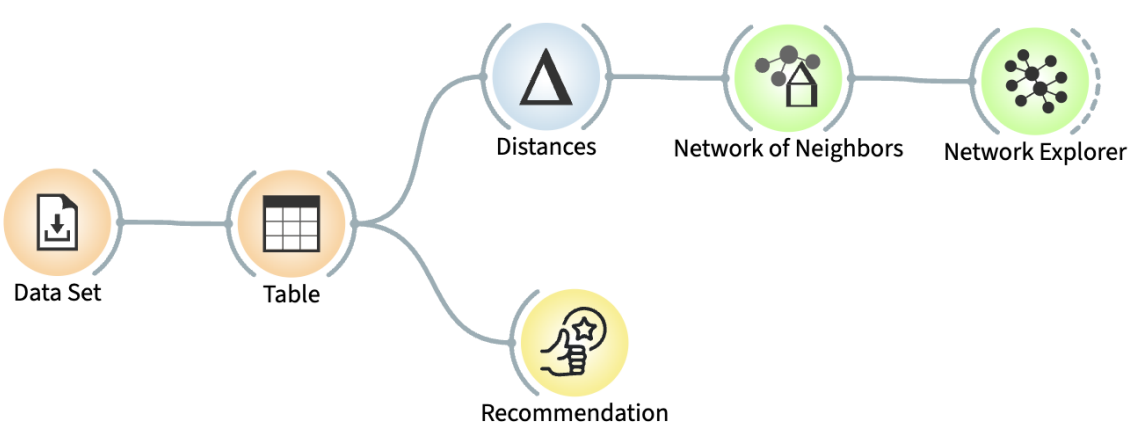

**Figure 8: The workflow behind the lesson about recommender systems. The Data Set widget loads the data, which are inspected in the Table widget. The Distance widget computes similarities between students, the Network of Neighbours widget constructs the network, and the Network Explorer widget visualizes it. Finally, the activity-specific Recommendations widget provides personalized recommendations for each student.**

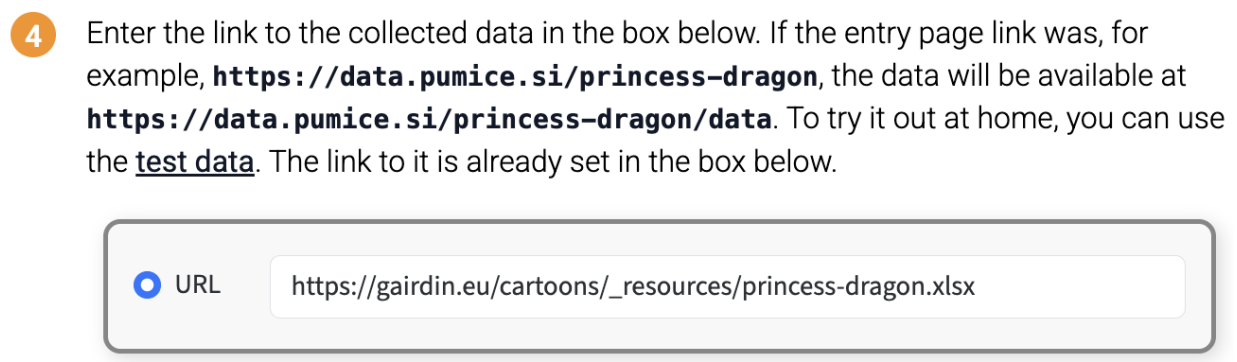

**Figure 9: The exposed part of the Data Set widget.**

Orange Lab solves these problems by only exposing the necessary functionality and by running on any device with a web browser. To keep lessons open and attractive, we start them with an unplugged activity that introduces the concepts on an intuitive level, and only then proceed with trying them on a computer.

We will present three such activities, with a focus on the latter part.

## 5.1 Recommender Systems: Favourite Cartoons

In the unplugged introductory activity, students select their favourite cartoons (or sitcoms, books, etc.). Working in groups, they construct a network in which each student is connected to the three peers with the most similar selections. The items chosen by these peers then serve as recommendations.

The teacher uses a dedicated web page to generate an event-specific URL through which students submit their selections. The collected data are then processed in Orange to replicate, at the level of the entire class, the procedure that students first carried out manually.

Figure 8 illustrates the underlying workflow. To conduct the lesson, teachers do not need to access the workflow; they only need access to a subset of widgets, each with a limited set of options. For example, in the Data Set widget, the only functionality that needs to be exposed is the field for entering the URL (Figure 9).

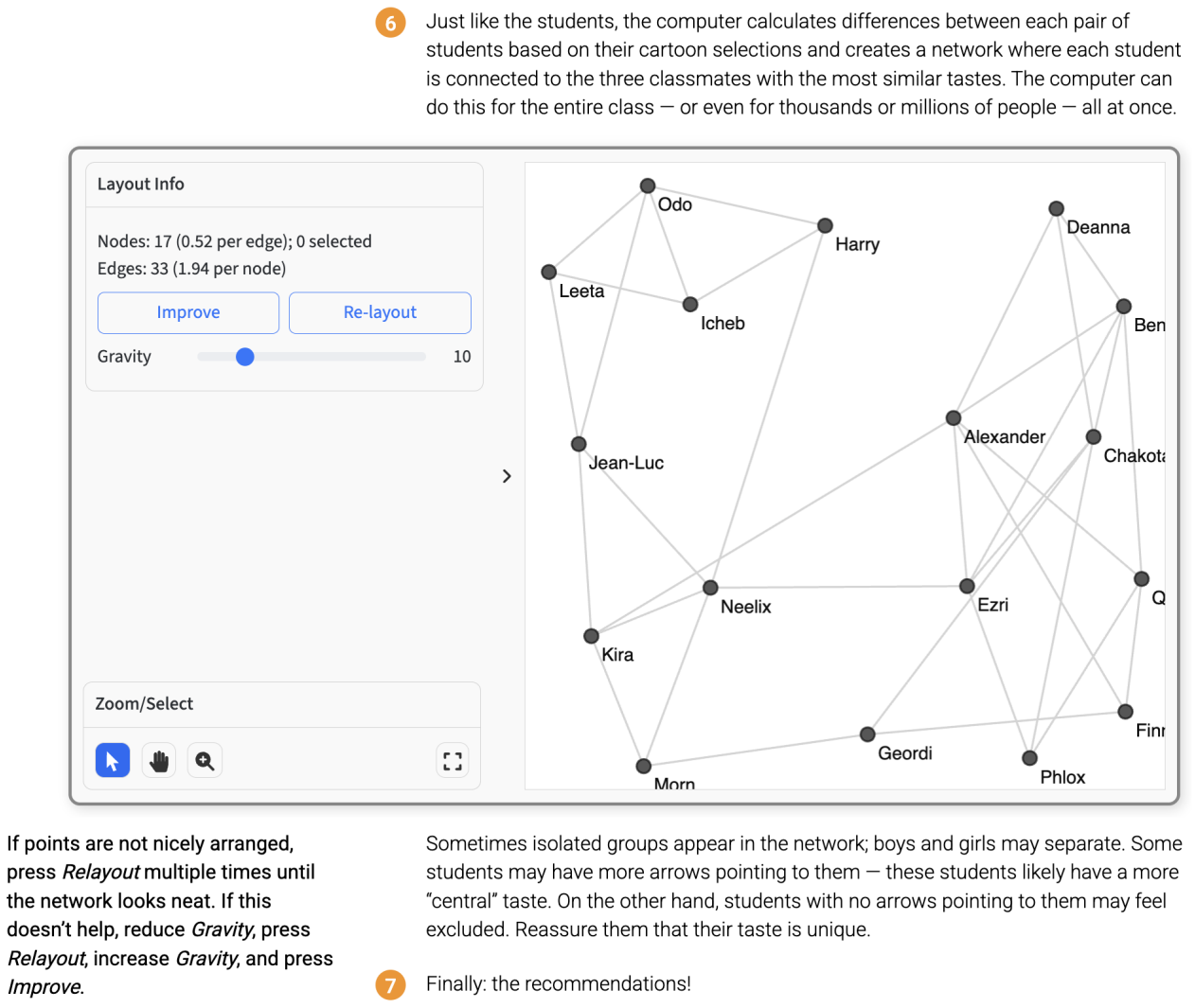

**Figure 10: The exposed part of the Network widget: we kept the options related to layout, but removed the unnecessary options for setting the properties of nodes and edges.**

Similarly, while the full Network Explorer widget includes many options, teachers typically require only the network visualization and, optionally, a few controls for adjusting its layout (Figure 10). In addition, we expose the Table widget, which allows both teachers and students to inspect the collected data, and the Recommendations widget.

These widgets are embedded within a web page that provides step-by-step instructions for conducting the activity, allowing teachers to follow the lesson without switching between multiple windows or tabs.

## 5.2 Classification: Gnomes and Animal Key

To introduce classification, we use a set of cards depicting gnomes whose professions can be inferred from their features. For example, gnomes holding shovels are either miners or gardeners; miners can be distinguished by also carrying a lamp. Other gnomes are either tailors (identified by a belt buckle) or masons. Additional features, such as shoe color or mouth shape, are irrelevant. In this unplugged activity, students are guided to discover the structure of classification trees.

Students then formulate a set of rules that enables them to determine a gnome's profession based on the characteristics of its house. While students are drawing the houses, the teacher generates an event-specific URL that provides a form for describing them. The collected data are then used by Orange Lab to induce a classification tree, which may – or may not – align with the rules proposed by the students.

This simpler workflow involves three widgets (Data, Tree, and Tree Viewer). We expose only the URL field of the Data widget and the right-hand panel of the Tree Viewer. For aesthetic reasons, the zoom controls for the tree are presented in a separate frame. For more advanced use, we also expose selected parameters of the tree induction algorithm (Figure 11).

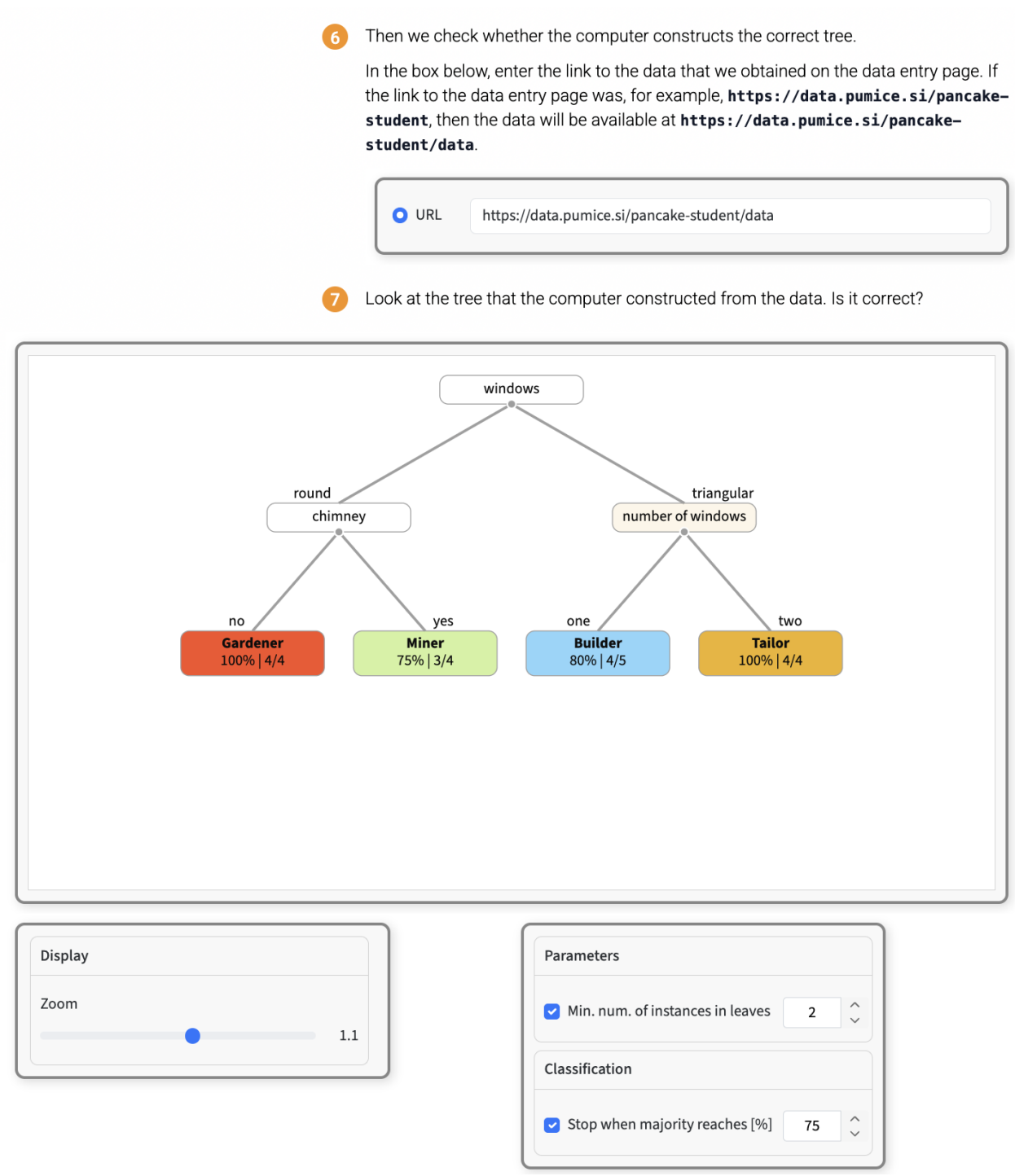

**Figure 11: Widgets for classifying gnome professions embedded in a web page. The teacher enters the data collection URL into the Data Set widget, after which Orange Lab constructs and displays the classification tree. The left-hand panel below the Tree Viewer is also part of the Tree Viewer but is presented separately for improved layout. The widget on the right can be used to demonstrate the effects of tree pruning, if desired.**

In a similar activity, students train a computer to distinguish between mammals, insects, birds, and other animal classes. The teacher edits a table in Google Sheets, adding rows with animals and their corresponding classes, and, as needed, new columns describing their features (e.g., "gives milk," "has feathers"). A File widget retrieves the data from this table and passes it to a Tree widget, which constructs the classification model, subsequently visualized in the Tree Viewer (Figure 12)

## 5.3 Clustering: Climate Zones

We introduce clustering by asking students to group cards containing climographs of 36 South Asian cities into four clusters based on similarity. They then describe each cluster in terms of shared characteristics (e.g., hot, dry summers and mild, rainy winters), assign a color to each group, and mark the corresponding cities on a map.

Students can then observe the same process on a web page, which displays the data and the cities, while also enabling interactive exploration. For example, the teacher or students can select a cluster to view its average climograph (Figure 5), or examine the climographs of individual or multiple cities by selecting them on the map.

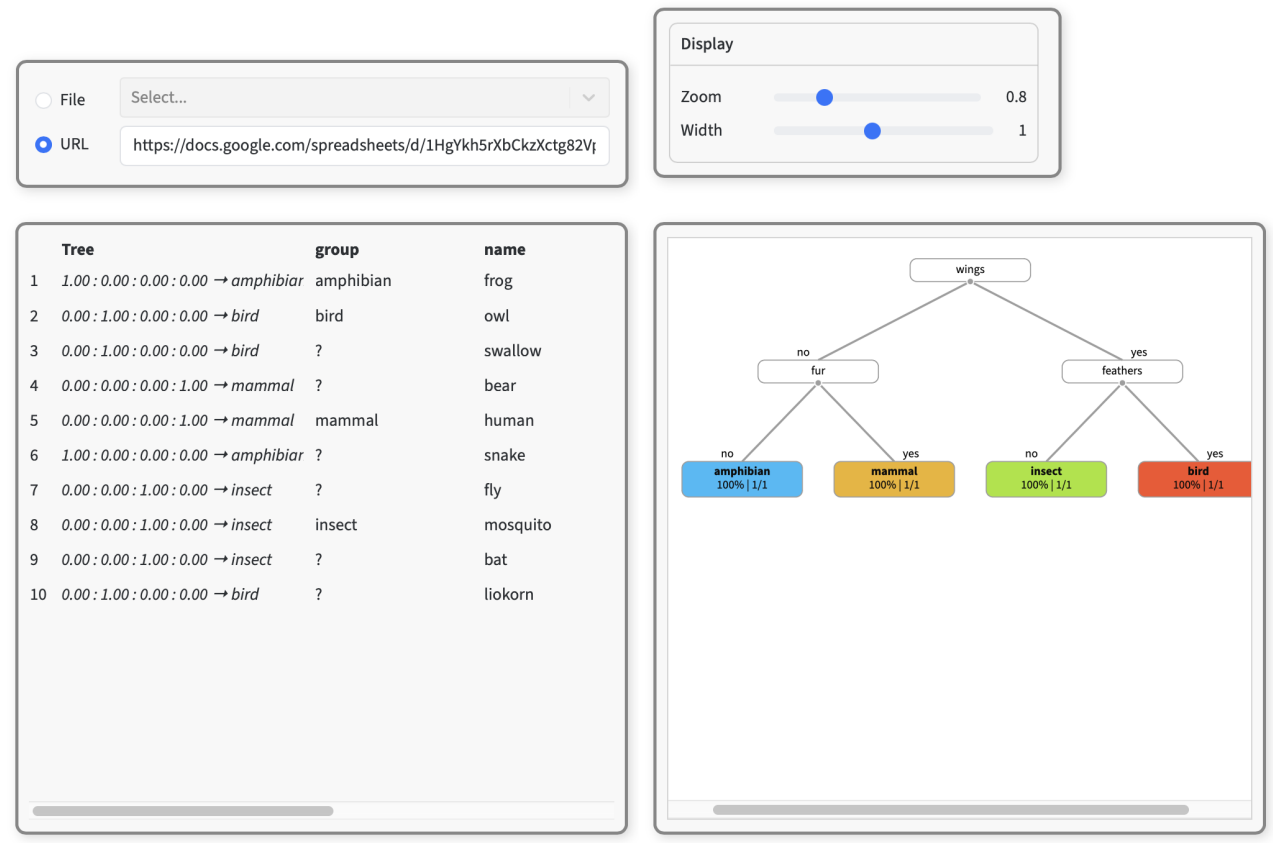

**Figure 12: Widgets for training a model to distinguish between classes of animals. This web page contains only the widgets, without any accompanying text. This design allows the teacher to place two browser windows side by side – editing the data in one while immediately observing the effects, such as the updated tree and predictions, in the other.**

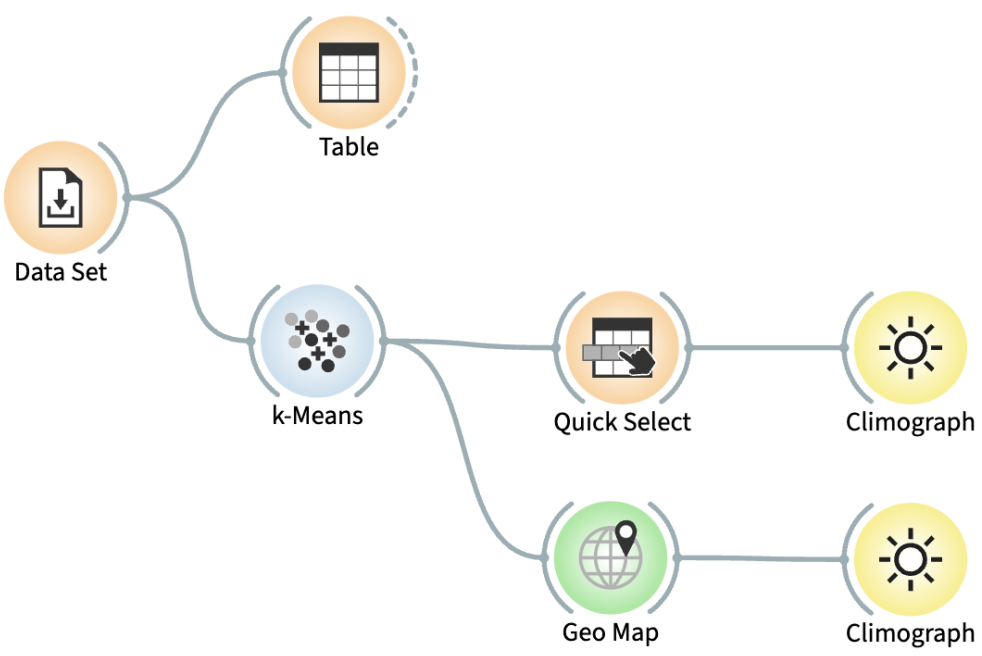

**Figure 13: Workflow for discovery of climate zones in South Asia using k-means clustering.**

This web page exposes all widgets from the corresponding workflow (Figure 13), while omitting most parameters and focusing on visualization. Teachers do not need to be aware of the underlying workflow to use these visualizations effectively. The arrangement of the exposed widgets, together with the accompanying instructions, makes the interactions intuitive – for instance, selecting a city on the map immediately updates the corresponding data shown in the adjacent graph (Figure 14).

These examples show how instead of exposing students to the full application, Orange Lab allows us to embed interactive widgets directly into web-based lessons. These widgets are customized (i.e. simplified) to match the pedagogical goals and provide guided interaction, allowing students to manipulate data, test hypotheses, and observe outcomes without requiring prior knowledge of the underlying tool.

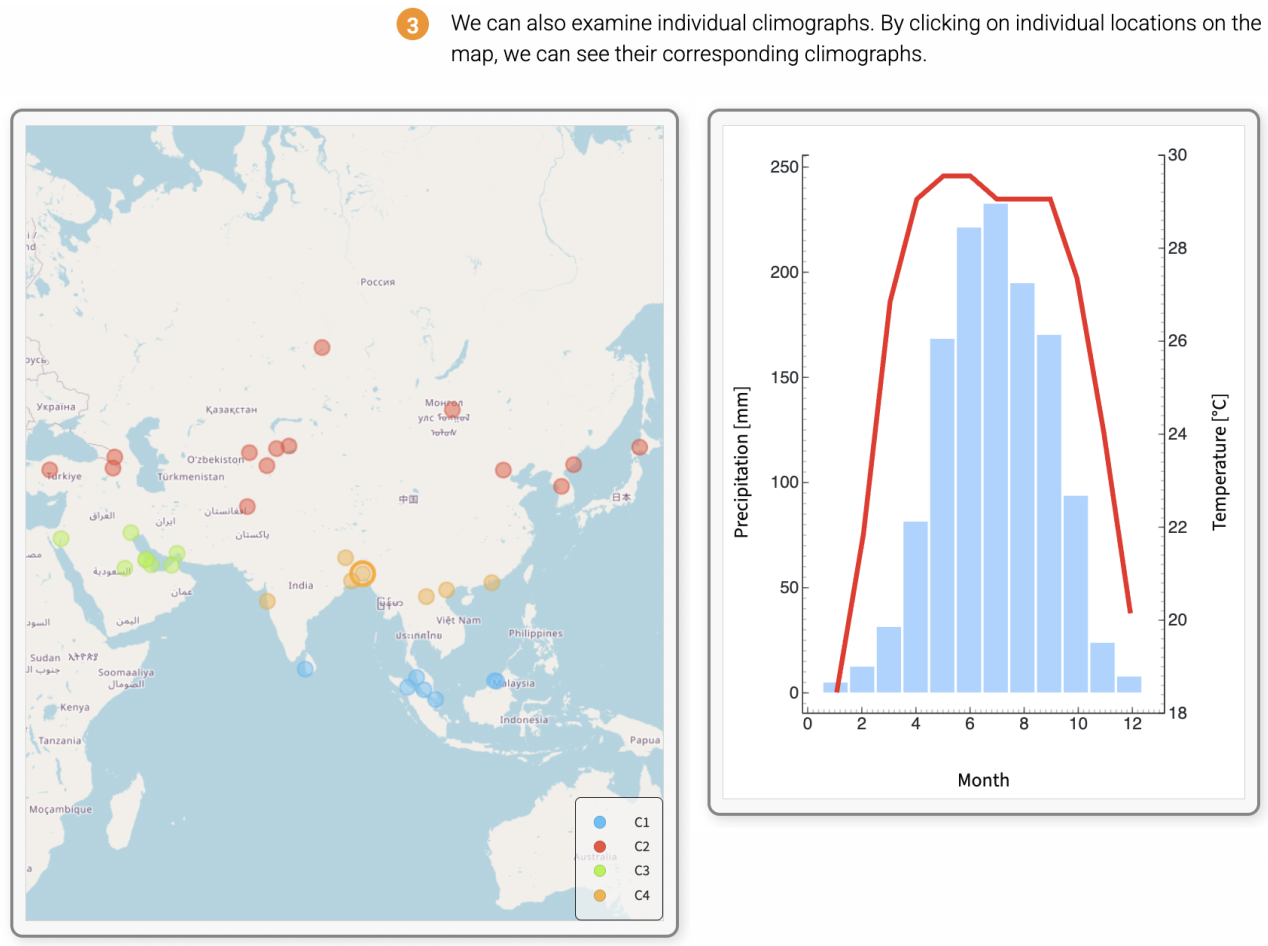

**Figure 14: A combination of Map widget and Climatograph. Selecting one (or more) cities in the former shows its (or their) average temperature and precipitation.**

Lessons like these were prepared in advance and provided to teachers to lower the barrier to introducing data literacy in the classroom. At the same time, more experienced teachers can use the same environment to design their own data analysis workflows, tailoring them to their specific teaching goals while shielding students from the technical complexities of the underlying system.

## 6 Discussion

Orange Lab is both a new data analytics tool and a shift in how data analysis workflows can be turned into interactive experiences and applications. By designing and selectively exposing components of the data analysis pipeline, Orange Lab can be used to create a workflow, use it interactively within the data analysis toolbox, or rewrap it within an entirely new web page, preserving the full functionality while exposing only the interface elements that matter to the user. In this way, Orange Lab supports the rapid creation of interactive applications without additional programming, as well as the reuse of existing workflows in new contexts.

The classroom deployments to support training of data literacy in schools show that Orange Lab's approach works in practice: teachers were able to use prepared materials with little effort, while still benefiting from the interactivity of the underlying workflows. For students, the cognitive load associated with learning a new tool was reduced, enabling them to focus on the conceptual understanding of data science principles.

The proposed approach has clear implications beyond the educational settings explored in our case studies. Namely, it provides the means for turning analytical workflows into lightweight, shareable applications that can be directly embedded into reports, dashboards, or any online content. This opens up new possibilities for communicating data-driven results, enabling audiences to interact with analyses rather than passively consume them.

Orange Lab is still in its infancy, and component exposition, while it has survived harsh testing in demanding educational environments, requires further study across various deployment contexts, particularly in industrial settings. Broader adoption of this concept will require improved support for the authoring and maintenance of embedded interfaces, especially in scenarios where applications and their multiple components must be coordinated or integrated with external systems. This includes more flexible mechanisms for controlling widget behavior from the host environment, as well as higher-level (e.g. drag-and-drop) authoring tools for composing interfaces, including dashboard-like layouts. Further opportunities lie in supporting automated, AI-based assistance during workflow construction and interface design, which could lower the barrier for less experienced users and speed up development.

## References

[1] Alteryx. [n. d.]. Alteryx. https://www.alteryx.com/. Accessed: 2026-03-24.
[2] Michael R Berthold, Nicolas Cebron, Fabian Dill, Thomas R Gabriel, Tobias Kötter, Thorsten Meinl, Peter Ohl, Kilian Thiel, and Bernd Wiswedel. 2009. KNIME-the Konstanz information miner: version 2.0 and beyond. *AcM SIGKDD explorations Newsletter* 11, 1 (2009), 26–31.
[3] Ekaba Bisong. 2019. Google colaboratory. In *Building machine learning and deep learning models on google cloud platform: a comprehensive guide for beginners*. Springer, 59–64.
[4] Anthony D Blaom, Franz Kiraly, Thibaut Lienart, Yiannis Simillides, Diego Arenas, and Sebastian J Vollmer. 2020. MLJ: A Julia package for composable machine learning. *arXiv preprint arXiv:2007.12285* (2020).
[5] Michael Bostock. [n. d.]. bqplot: Plotting library for Jupyter notebooks. https://github.com/bqplot/bqplot. Accessed: 2026-03-31.
[6] Michael Bostock, Vadim Ogievetsky, and Jeffrey Heer. 2011. $D^3$ data-driven documents. *IEEE transactions on visualization and computer graphics* 17, 12 (2011), 2301–2309.
[7] Michelle Carney, Barron Webster, Irene Alvarado, Kyle Phillips, Noura Howell, Jordan Griffith, Jonas Jongejan, Amit Pitaru, and Alexander Chen. 2020. Teachable machine: Approachable web-based tool for exploring machine learning classification. In *Extended abstracts of the 2020 CHI conference on human factors in computing systems*. 1–8.
[8] Project Jupyter Contributors. [n. d.]. ipywidgets: Interactive widgets for the Jupyter Notebook. https://github.com/jupyter-widgets/ipywidgets. Accessed: 2026-03-31.
[9] Voilà contributors. [n. d.]. Voilà: Rendering of live Jupyter notebooks with interactive widgets. https://github.com/voila-dashboards/voila. Accessed: 2026-03-31.
[10] Dataiku. [n. d.]. Dataiku. https://www.dataiku.com/. Accessed: 2026-03-24.
[11] Janez Demšar, Tomaž Curk, Aleš Erjavec, Črt Gorup, Tomaž Hočevar, Mitar Milutinovič, Martin Možina, Matija Polajnar, Marko Toplak, Anže Starič, et al. 2013. Orange: data mining toolbox in Python. *the Journal of machine Learning research* 14, 1 (2013), 2349–2353.
[12] Jules Françoise, Baptiste Caramiaux, and Téo Sanchez. 2021. Marcelle: composing interactive machine learning workflows and interfaces. In *The 34th Annual ACM symposium on user interface software and technology*. 39–53.
[13] Zhiyi Gao, Yonghong Hou, Yan Liu, and Xiangyu Li. 2021. MetaFlow: a meta-learning-based network for optical flow estimation. *Journal of Electronic Imaging* 30, 3 (2021), 033029–033029.
[14] Mark Hall, Eibe Frank, Geoffrey Holmes, Bernhard Pfahringer, Peter Reutemann, and Ian H Witten. 2009. The WEKA data mining software: an update. *ACM SIGKDD explorations newsletter* 11, 1 (2009), 10–18.
[15] Markus Hofmann and Ralf Klinkenberg. 2016. *RapidMiner: Data mining use cases and business analytics applications*. CRC Press.
[16] Thomas KLUYVER[a1], Benjamin RAGAN-KELLEYb[1], Fernando Pérez, Brian Granger, Matthias Bussonnier, Jonathan Frederic, and Kyle Kelley. 2016. Jupyter Notebooks-a publishing format for reproducible computational workflows. In *Positioning and power in academic publishing: players, agents and agendas: proceedings of the 20th International Conference on Electronic Publishing*. IOS press, 87.
[17] Vijay Kotu and Bala Deshpande. 2014. *Predictive analytics and data mining: concepts and practice with rapidminer*. Morgan Kaufmann.
[18] TMD Technology Limited. [n. d.]. Embeddable. https://embeddable.com/. Accessed: 2026-03-24.
[19] Randall Matignon. 2007. *Data mining using SAS enterprise miner*. John wiley & sons.

[20] Microsoft. [n. d.]. Azure Machine Learning Designer. https://learn.microsoft.com/en-us/azure/machine-learning/concept-designer. Accessed: 2026-03-24.
[21] Microsoft. [n. d.]. Lobe: Machine Learning Made Easy. https://github.com/lobe. Accessed: 2026-03-24.
[22] Viktoriya Olari and Ralf Romeike. 2025. Teaching Data Concepts and Practices in Secondary School Education on Artificial Intelligence: Approaches, Mechanisms, and Emerging Local Theories. In *Proceedings of the 25th Koli Calling International Conference on Computing Education Research*. 1–12.
[23] Inc. SageMath. [n. d.]. CoCalc. https://cocalc.com/. Accessed: 2026-03-31.
[24] Salesforce. [n. d.]. Tableau. https://www.tableau.com/. Accessed: 2026-03-24.
[25] Arvind Satyanarayan, Dominik Moritz, Kanit Wongsuphasawat, and Jeffrey Heer. 2016. Vega-lite: A grammar of interactive graphics. *IEEE transactions on visualization and computer graphics* 23, 1 (2016), 341–350.
[26] Sisense. [n. d.]. Sisense. https://www.sisense.com/. Accessed: 2026-03-24.
[27] MLJAR sp. [n. d.]. Mercury: Create web apps from Python notebooks. https://github.com/mljar/mercury. Accessed: 2026-03-31.
[28] Tilo Wendler and Sören Gröttrup. 2016. *Data mining with SPSS modeler: theory, exercises and solutions*. Springer.
[29] Bowen Yu and Claudio T Silva. 2016. VisFlow-web-based visualization framework for tabular data with a subset flow model. *IEEE transactions on visualization and computer graphics* 23, 1 (2016), 251–260.